\documentclass[10pt,twocolumn,letterpaper]{article}

\usepackage{iccv}
\usepackage{times}
\usepackage{epsfig}
\usepackage{graphicx}
\usepackage{amsmath}
\usepackage{amssymb}

\usepackage{algorithmic}
\usepackage{algorithm}
\usepackage{times}
\usepackage{epsfig}
\usepackage{graphicx}
\usepackage{amsmath}
\usepackage{amssymb}
\usepackage{mathtools}
\usepackage{amsmath,amssymb,amsfonts}
\usepackage{amsthm}
\usepackage{titling}
\usepackage{multicol}
\usepackage{graphicx}
\usepackage{textcomp}
\usepackage{xcolor}
\usepackage{bbm}
\usepackage[pagebackref=true,breaklinks=true,letterpaper=true,colorlinks,bookmarks=false]{hyperref}

\iccvfinalcopy 

\def\iccvPaperID{5960} 

\ificcvfinal\fi

\begin{document}

\title{Shared and Private VAEs with Generative Replay for Continual Learning}

\author{Subhankar Ghosh \\
Indian Institute of Science\\
Bengaluru, India\\
{\tt\small subhankarg@alum.iisc.ac.in}
}
\date{}
\maketitle
\ificcvfinal\thispagestyle{empty}\fi

\begin{abstract}
   Continual learning tries to learn new tasks without forgetting previously learned ones. In reality, most of the existing artificial neural network(ANN) models fail, while humans do the same by remembering previous works throughout their life. Although simply storing all past data can alleviate the problem, it needs large memory and often infeasible in real-world applications where last data access is limited. We hypothesize that the model that learns to solve each task continually has some task-specific properties and some task-invariant characteristics. We propose a hybrid continual learning model that is more suitable in real case scenarios to address the issues that has a task-invariant shared variational autoencoder and T task-specific variational autoencoders. Our model combines generative replay and architectural growth to prevent catastrophic forgetting. We show our hybrid model effectively avoids forgetting and achieves state-of-the-art results on visual continual learning benchmarks such as MNIST, Permuted MNIST(QMNIST), CIFAR100, and miniImageNet datasets. We discuss results on a few more datasets, such as SVHN, Fashion-MNIST, EMNIST, and CIFAR10. Our code is available at \textcolor{blue}{{\url{https://github.com/DVAEsCL/DVAEsCL}}}.
\end{abstract}

\section{Introduction}
\begin{figure}
    \centering
    \includegraphics[width = 8cm, height = 5cm]{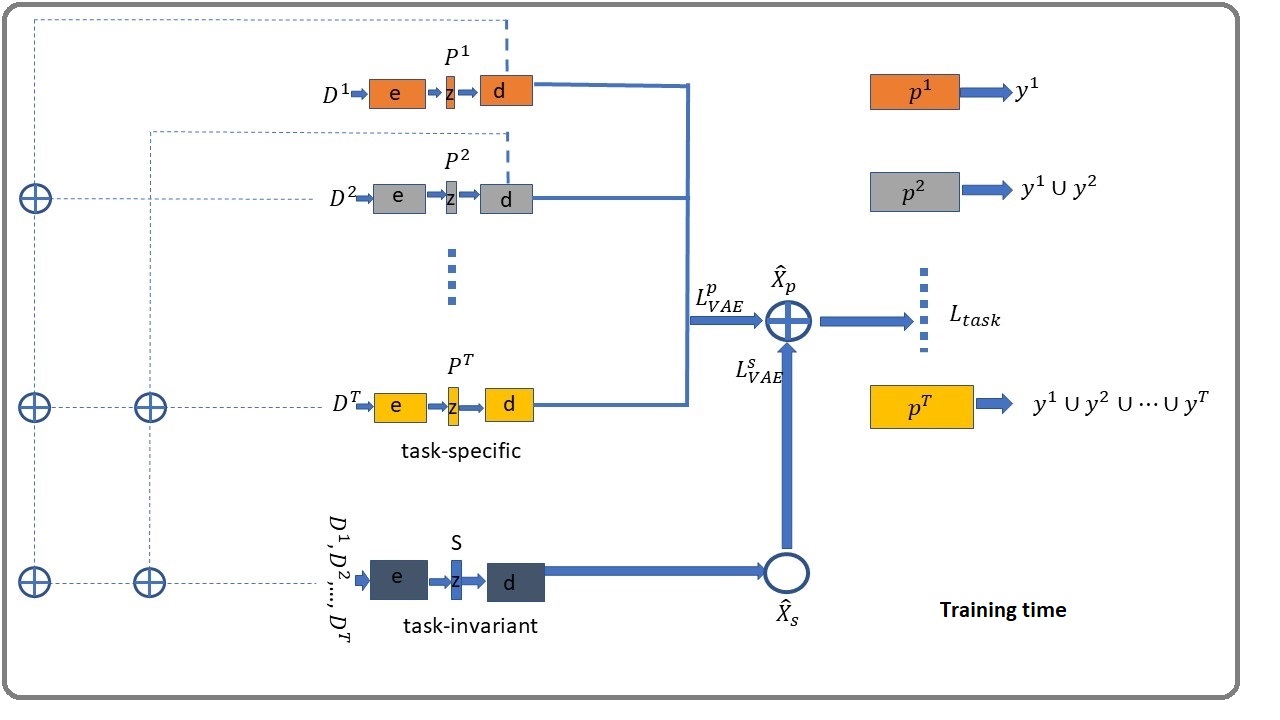}
    \caption{ is our model at training time. Architecture growth occurs at the arrival of $t^{th}$ task by adding a task-specific VAE denoted as $P^t$, and a task-specific ANN denoted as $p^t$. To prevent forgetting, private VAEs are stored for each task. The first private VAE gets trained using real data from the first task. The second private VAE is trained using real data from the second task and synthesized data corresponding to the first task's classes generated from the first private decoder. Similarly, the $t^{th}$  private VAE sees real data from the $t^{th}$ task, synthesized data from the previous private decoders corresponding to their tasks' classes during training. A shared VAE that is less prone to forgetting yet is also retrained with a small number of generative replay. The plus(+) sign indicates concatenation.
}
    \label{fig:f1}
\end{figure}
Humans and animals are capable of continually learning and updating knowledge throughout their lifetime. The ability to accommodate new experiences while retaining previously known knowledge helps to build reusable artificial intelligent systems. Current ANNs achieve impressive performance on many machine learning problems like image classification, object detection, and natural language processing but fail to remember previous knowledge due to a phenomenon called catastrophic forgetting\cite{a1} when trained for new tasks. We want our artificial learning agents to have the ability to solve many tasks sequentially under different conditions by developing task-invariant and task-specific skills that enable them to adapt while avoiding forgetting using generative replay quickly. 
Several approaches have been proposed over the years to alleviate the catastrophic forgetting. The first approach involves dynamically increasing the network's capacity to learn new tasks\cite{a2, a3}. The second approach uses regularizers that make the number of parameters remain constant during sequential learning\cite{a4, a5, a6, a7, a8}. However, these approaches are not feasible if each task needs a large memory. Another line of approach is to rely either on experience replay\cite{a9, a10, a11} or generative replay\cite{a12, a13, a14} by storing real data from previous tasks or train generative models, respectively.

\begin{figure}
    \centering
    \includegraphics[width = 8cm, height = 5cm]{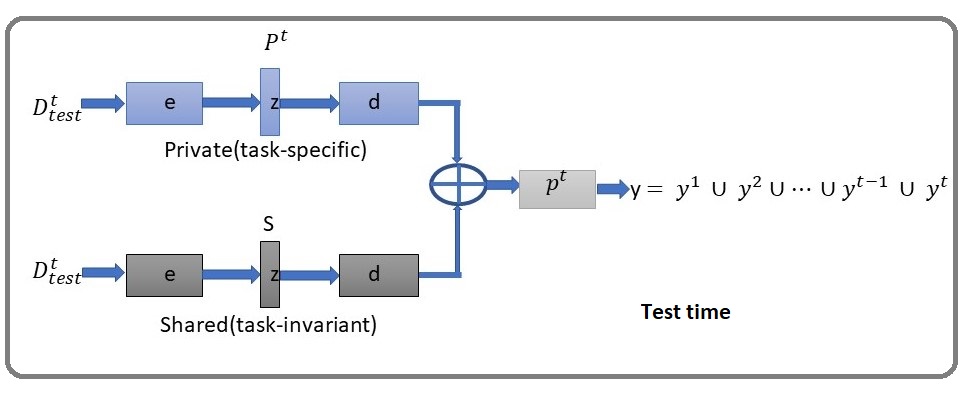}
    \caption{ is our model during test time at $t^{th}$ task. The model gets data corresponding to all classes till the $t^{th}$ task and predicts those classes 
}
    \label{fig:f2}
\end{figure}
We propose a novel continual learning method in which the network is dynamic, uses generative replay, and grows in size with each task. A disjoint space representation composes task-invariant or shared space(fixed size) trained for all tasks, and task-specific or private space grows with each task. Our approach is motivated by the fact that our human brain structure is complex and contains billions of neurons\cite{a15}. We may need to move towards a more complex neural network structure in the coming future to make artificial systems solve the works humans can do nowadays. Continual learning can be applied to various practical situations involving privacy issues. The main contributions of these work are summarized as follows:
\begin{itemize}
\item We develop a structure-based and generative replay-based model using (T + 1) numbers of conditional variational autoencoders. T is the number of tasks the model has to solve.
\item We show the private and shared modules' training is different from\cite{a16} though that paper inspires our network's architecture.
\item We present results for some datasets and show our model achieves state-of-the-art performance on all the datasets.
\end{itemize}
\section{Related Work}
\subsection{Continual Learning}
The existing continual learning approaches can be divided into architecture-based, regularization-based, and rehearsal-based strategies.
\subsection*{Architecture-based Methods}
The first approach to prevent catastrophic forgetting does modification in the network's structure by growing a module for each task either physically or logically\cite{a2, a7}. These methods attempt to localize inference to a subset of the network such as columns\cite{a17}, neorons\cite{a18, a19}, a mask over parameters\cite{a20, a21}. The performance of learned tasks is preserved by storing the learned modules while accommodating new tasks by augmenting the network with new modules. PNNs\cite{a22} statically grow the architecture, are immune to forgetting, and utilize prior knowledge via lateral connections to previously learned features. Reference\cite{a19} proposed a dynamically expandable network(DEN) that can dynamically decide its network's capacity as it gets trained on a sequence of tasks. These methods impose a computational cost in continual learning where many tasks need to be learned, and fixed capacity memory can not be considered. 
\subsection*{Regularization-based Methods}
The second family of this field is based on regularization. They estimate the importance of a network's parameters and penalize those weights while switching from one task to another task. There are many existing approaches for penalizing the weights. One of the methods is the elastic weight consolidation[EWC]\cite{a23}, where important parameters have the highest in terms of the Fisher information matrix. In reference\cite{a8}, the weights are computed online and kept track of how much the loss changes due to change in specific weights and accommodate this information during training. Reference\cite{a4} focuses on the change on the activation instead of considering the loss's change. This way, parameter importance is learned in an unsupervised manner. Despite the success gained by these methods, they are often limited by the number of tasks.
\subsection*{Rehearsal-based Methods}
The final family of methods of this domain to mitigate forgetting is rehearsal-based. Existing approaches use two strategies: either store few samples per class from previous tasks or train a generative model like GAN\cite{a24} or VAE\cite{a25} or both to sample synthetic data from previously learned distributions. The iCaRL\cite{a11} stores a subset of real data(exemplars). For a constant memory budget, the number of data stored per learned class decreases as the number of tasks increases, so the models' performance decay. Reference\cite{a26} proposed two losses called: the less-forget constraint and inter-class separation to prevent forgetting. The less-forget loss minimizes the cosine distance between the features extracted from the old and new models. References\cite{a27, a28} introduce a bias-correction layer to correct the original fully-connected layer's output to address the data imbalance between the old and new categories. A recent study on tiny episodic memories in continual learning are GEM\cite{a29}, A-GEM\cite{a30}, MER\cite{a31}, and ER-RES\cite{a32}. 

The second strategy in this family does not store any data but generates synthetic data using generative models. Reference\cite{a13} used generative replay with an unconditional GAN, where an auxiliary classifier needs to determine which classes the generated samples belong. Reference\cite{a14} is an improved version of\cite{a13}, where they used class-conditional GAN to synthesize data. Reference\cite{a33} used a generative autoencoder for replay. Synthetic data for previous tasks are generated based on the mean and covariance matrix using the encoder's class statistics. The major limitation of these approaches is the assumption of a Gaussian distribution of the data. 
\subsection{Space Factorization}

In machine learning, multi-view learning is more effective, more promising, and has a better generalization ability than single-view learning\cite{a34}. The approaches to tackle multi-view learning aim at either maximizing the mutual agreement on different views of the data or focus on gaining sub-space shared by many views by assuming that the input views are synthesized from that subspace using clustering\cite{a35}, Gaussian processes\cite{a36}, etc. So, the concept of factorizing the space into shared and private sub-spaces has been explored\cite{a37}. In this paper, we factorize the data space into two parts: shared and private. 
\section{Shared and Private VAEs with Generative Replay for Continual Learning}
    We study the problem of learning a sequence of T data distributions denoted as $D = \{D^1, D^2, ..., D^T\}$, where $D^t = \{(X^t_i, Y^t_i, T^t_i)^{n_t}_{i = 1}\}$ is the data distribution for the task t with $n_t$ sample tuples of input($X^t \in \mathcal{X}$), target label ($Y^t \in \mathcal{Y}$), and task label($T^t \in \mathcal{T}$). The goal is to learn a sequential function, $f_\theta: D^t \rightarrow \hat{\mathcal{Y}}^t$, for each task, where $\hat{\mathcal{Y}}^t$ are the predicted labels corresponding to $t^{th}$ task. $f_\theta \in (f_S \cup f_P \cup f_p)$, where $f_S: D^t \rightarrow \hat{\mathcal{X}}_S, f_P:D^t \rightarrow \hat{\mathcal{X}}_P$, and $f_p:\hat{\mathcal{X}}_S \cup \hat{\mathcal{X}}_P \rightarrow \hat{\mathcal{Y}}^t$. We try to achieve our goal by training two separate modules: private and shared, to mitigate forgetting of prior knowledge. The model prevents catastrophic forgetting in shared and private spaces separately and begins learning $f_{\theta}^t$ where $\theta \in (\theta_S, \theta_P, \theta_p)$ as mapping function from $\mathcal{D}^t$ to $\mathcal{Y}^t$. We use some n samples per class to be synthesized prior to $t^{th}$ task and accumulate the generated data to the current task($t^{th}$) to train the model. During training the model with $t^{th}$ task:
\begin{center}
    $X^{t} \leftarrow X^{t}\cup \hat{\mathcal{X}}^{1:(t-1)}_p$
\end{center}
The cross-entropy loss function for the $f^t_{\theta}$ mapping corresponds to:
\begin{equation}
    \begin{split}
    L_{task = t}(f^t_{\theta}, D^t) = -\mathop{\mathbb{E}}_{(\mathcal{X}^t, \mathcal{Y}^t) \sim D^t}\\
    \left[\sum_{c = 1}^C\mathbbm{1}_{(c = \mathcal{Y}^t)}log(\sigma(f_{\theta}^t(\mathcal{X}^t, \mathcal{Y}^t)))\right]
    \end{split}
\end{equation}
Where $\sigma$ is the softmax function, in learning a sequence of tasks, an ideal $f^t_{\theta}$ maps the input images $X^t$ to their predicted labels $\hat{\mathcal{Y}}^t$.

\subsection*{Variational autoencoders(VAEs)}
Autoencoders can effectively learn input space and representation\cite{a38, a39}. A VAE is a generative model that follows an encoder-latent vector-decode architecture of classical autoencoder, which places a prior distribution on the input space and uses an expected lower bound to optimize the learned posterior. Conditional VAE is an improved version of the VAE, where data are fed to network with class properties such as labels. The variational autoencoderis a fundamental building block of our approach. Variational distribution tries to find a true conditional probability distribution over the latent space z through minimizing their distance using a variational lower bound limit. The loss function for a VAE is:
\begin{multline}
    L_{VAE} = \mathop{\mathbb{E}}_{q_{\phi}({z}|{x})}\left[log(p_\theta({x}|{z}))\right] - D_{KL}(q_\phi({z}|{x})\parallel p_{\theta(z)})
\end{multline}
Where the first one is the reconstruction loss, and the second term is the KL divergence between $q({z}|{x})$ and p(z). The encoder predicts $\mu$ and $\sum$ such that $q_{\phi}({z}|{x}) = \mathcal{N}(\mu, \sum)$, from which a latent vector is synthesized via reparametrization process.

The final objective function of our approach for the $t^{th}$ task is: 
\begin{multline}
L^{(t)} = \lambda_1 L_{task} + \lambda_2 L^S_{VAE} + \lambda_3 L^P_{VAE}
\end{multline}
Where, $\lambda_1, \lambda_2,$ and $\lambda_3$ are regularizer constants to control the effect of each loss component. The working algorithm of this model is presented in Algorithm \ref{alg:1}. 
\subsection{Avoid forgetting}
Catastrophic forgetting occurs because of the imbalance in the data between previous and new classes that creates a bias in the network towards the current ones during training, and models almost forget previous knowledge. One of our approach's insights is to decouple the single space learned for all tasks continually into two parts: shared and private sub-spaces. Another approach is the generative replay from previous private modules that concatenated into the current task's data during training the model with the current task to avoid forgetting. The first private module sees real data of the first task; the second private  VAE sees real data of the second task and synthetic data of the first task. Similarly, the third private module gets real data of the third task and synthetic data of the first and second tasks generated from the first and second private decoders, respectively, during training the model with the third task. Whatever data individual private module sees during training, the only shared module gets trained by them. It goes like this till the $T^{th}$ task. 
\subsection{Evaluation Matrices}
We estimate the resulting model on all previous tasks similar to\cite{a29, a41} after training for each new task. We use ACC as the average test classification accuracy across all classes for continual learning to measure our model's performance. To measure forgetting, we calculate backward transfer, BWT that calculates how much learning new tasks have influenced previous tasks' performance. While $BWT<0$ indicates catastrophic forgetting and $BWT>0$, learning new tasks has helped improve performance on previous tasks. 
\begin{equation}
    BWT = \frac{1}{T-1}\sum^{T-1}_{t = 1}\left[R_{T,t} - R_{t, t}\right]
\end{equation}
\begin{equation}
    ACC = \frac{1}{T}\sum^{T}_{t = 1}R_{t, t}
\end{equation}
ACC is the mean classification accuracy across all tasks.
Where $R_{j,i}$ is the test classification accuracy on task i after sequentially finishing learning the $j^{th}$ task.

\begin{algorithm}
\caption{Continual Learning}
\label{alg:1}
\textbf{Input}: $(\mathcal{X}, \mathcal{Y}) \sim D^{all}$\\
\textbf{Parameters}: $\theta \in (\theta_S \cup \theta_P \cup \theta_p)$\\
\textbf{Output}: $\mathcal{Y}$
\begin{algorithmic}[1] 

\STATE $D_{gen} \leftarrow \{\}$
\FOR{t $\leftarrow$ 1 to T}
  \FOR{e $\leftarrow$ 1 to epochs}
    
      \STATE Compute $L_{task}$ using $(\mathcal{X}^t, \mathcal{Y}^t) \in D^t$ 
      \STATE Compute $L_{VAE}^S$ for the shared module using $(\mathcal{X}^t, \mathcal{Y}^t)\in D^t$
      \STATE Compute $L_{VAE}^P$ for the $t^{th}$ private module using $(\mathcal{X}^t, \mathcal{Y}^t)\in D^t$
      \STATE $L^{(t)} = \lambda_1 L_{task} + \lambda_2 L^S_{VAE} + \lambda_3 L^P_{VAE}$
      \STATE $\theta^{'} \leftarrow \theta - \alpha \nabla L^{(t)}$

  \ENDFOR
  \STATE accuracy $\leftarrow$ function TEST($\mathcal{X}^t_{test}$, $\mathcal{Y}^t_{test}$, t)

  \FOR{c $\leftarrow$ 1 to C} 
  \STATE C is the replay classes.
    \FOR{i $\leftarrow$ 1 to n} 
    \STATE n is the number of samples to be generated per class for the experience replay.
    \STATE $\hat{\mathcal{X}}^i_p \sim D^{1:t}_{gen}$
      
    \ENDFOR
  \ENDFOR
  \STATE $\mathcal{X}^{t+1} \leftarrow \mathcal{X}^{t+1}\cup \hat{\mathcal{X}}_p$

\ENDFOR
\end{algorithmic}
\end{algorithm}
\section{Experiments}
This section consists of the datasets and baselines we used in our experiments and the implementation details.
\subsection*{Datasets}
We perform our approach on the commonly used benchmark datasets for T-Split continual learning, where the entire dataset is divided into T tasks. We use \textbf{5-Split MNIST} and \textbf{Permuted MNIST(QMNIST)}\cite{a42} previously used in\cite{a37, a41, a43, a44}, \textbf{20-Split miniImageNet}[45] used in\cite{a32, a37, a46}, \textbf{20-Split CIFAR100}\cite{a47} used in\cite{a29, a30, a37, a44}. We also perform our experiments on \textbf{5-Split SVHN}\cite{a48}, \textbf{5-Split CIFAR10}\cite{a47}, \textbf{5-Split Fashion-MNSIT}\cite{a50} and \textbf{13-Split not-MNIST(EMNIST)}\cite{a49}. The datasets' statistics are given in Table \ref{table:t1}. We have not used any data augmentation techniques.\\
\begin{table}
\begin{tabular}{ |p{1.5cm}|p{1cm}|p{0.8cm}|p{1.7cm}|p{0.85cm}|p{0.82cm}|  }
 \hline
 Dataset & \#Classes & \#Tasks & Input Size & \#Train Data(k) & \#Test Data(k)\\
 \hline
 MNIST   & 10    &5&$1\times28\times28$& 50& 10\\
 QMNIST&  10 &5&$1\times28\times28$& 60&50\\
 EMNIST&26&13&$1\times28\times28$& 100&12\\
 F-MNIST&10&5&$1\times28\times28$& 60&10\\
 CIFAR10& 10&5&$3\times32\times32$& 50&10\\
 CIFAR100& 100&20&$3\times32\times32$& 50&10\\
 SVHN&  10&5&$3\times32\times32$& 61&12\\
 mImageNet&  100&20&$3\times84\times84$& 50&10\\

 \hline
\end{tabular}
\caption{Statistics of the datasets. Where, EMNIST = not-MNIST, QMNIST = Permuted MNIST, F-MNIST = Fashio-MNIST, and mImageNet = miniImageNet.}
\label{table:t1}
\end{table}
 
\begin{table}[hbt!]

\begin{tabular}{ |p{1.9cm}|p{1cm}|p{1.8cm}|p{2.5cm}|}
 \hline
 Dataset & Encoder & Decoder & Preceptron \\
 \hline
 MNIST   & 4 CL    &4 DCL& 4 CL and 1 FC\\
 \hline
 QMNIST& 4 CL    &4 DCL& 4 CL and 1 FC\\
 \hline
 EMNIST& 4 CL    &4 DCL& 4 CL and 1 FC\\
 \hline
 Fashion-MNIST& 4 CL    &4 DCL& 4 CL and 1 FC\\
 \hline
 CIFAR10& 4 CL    &4 DCL&  2 CL and 2 FC\\
 \hline
 CIFAR100& 4 CL    &4 DCL& 2 CL and 1 FC\\
 \hline
 SVHN& 4 CL    &4 DCL& 2 CL and 2 FC\\
 \hline
 miniImageNet& 5 CL    &5 DCL& 4 CL and 1 FC\\

 \hline
\end{tabular}
\caption{The information of the networks used in experiments(CL = Convolution layers, DCL = Deconvolution layers, FC = Fully connected layers).}
\label{table:t2}
\end{table}
\subsection*{Baselines}
We compare with state-of-the-art approaches, including elastic weight consolidation(EWC)\cite{a23}, Progressive neural networks(PNNs)\cite{a22}, Hard Attention Mask(HAT)\cite{a21}, and ACL\cite{a37} using implementations given by\cite{a37} unless otherwise stated. We compare a few memory-based methods A-GEM\cite{a30}, GEM\cite{a29}, ER-RES\cite{a32} for MNIST, permuted MNIST(QMNIST), 20-Split CIFAR100, and 20-Split miniImageNet. We depend on the implementation provided by\cite{a37}. On Permuted MNIST results for SI\cite{a44} are taken from\cite{a21}, for VCL\cite{a43}, those are taken from\cite{a37}, and for uncertainty-based CL in Bayesian framework(UCB)\cite{a41} are taken from the actual paper. 
\begin{table}[hbt!]

\begin{tabular}{ |p{2.7cm}|p{1.7cm}|p{1.8cm}|}
 \hline
 Dataset & z dimension & \#parameters \\
 \hline
 MNIST   & 108    & 199019\\
 QMNIST& 108    & 199019\\
 EMNIST& 108    & 562939\\
 Fashion-MNIST& 108    & 199019\\
 CIFAR10& 192    & 3388428\\
 CIFAR100& 192    & 13790903\\
 SVHN& 192    & 3388428\\
 miniImageNet& 96    & 4107152\\

 \hline
\end{tabular}
\caption{The second column gives the dimension of latent space for shared and private VAEs for all datasets. The third column shows the number of parameters required to train each dataset.}
\end{table}
\label{table:t3}
\subsection*{Implementation Details}
The information of the networks used in our model is given in Table \ref{table:t2} for each dataset. For a dataset, the architectures of shared and private modules are the same. We take PyTorch as our working framework. We train each of all datasets for 50 epochs and evaluate our model's performance at the $25^{th}$ and $50^{th}$ epochs. We perform experiments using 4, 20, and 100 synthetic samples per class during continual training. The Adam optimizer\cite{a40} has been used for all experiments, and the learning rate for the model is 0.0001. The dimension of the latent variables z and the number of parameters used for each dataset are given in Table 3. We take $\lambda_1 = \lambda_2 = \lambda_3 = 1$ at the loss function. We use a Tesla V100 gpu in all our experiments.
\section{Results and Discussion}
	In the first set of experiments, we measure ACC, BWT, and the memory used by our method and compare it against state-of-the-art methods on 20-Split miniImageNet, 20-Split CIFAR100, 5-Split Permuted MNIST, and 5-Split MNIST. Next, we demonstrate the experiments on sequentially learning single datasets such as SVHN, Fashion-MNIST, EMNIST, and CIFAR10.

\subsection{Performance on 20-Split miniImageNet Dataset}
\begin{figure}[hbt!]
\centering
    \includegraphics[width = 8cm, height = 5cm]{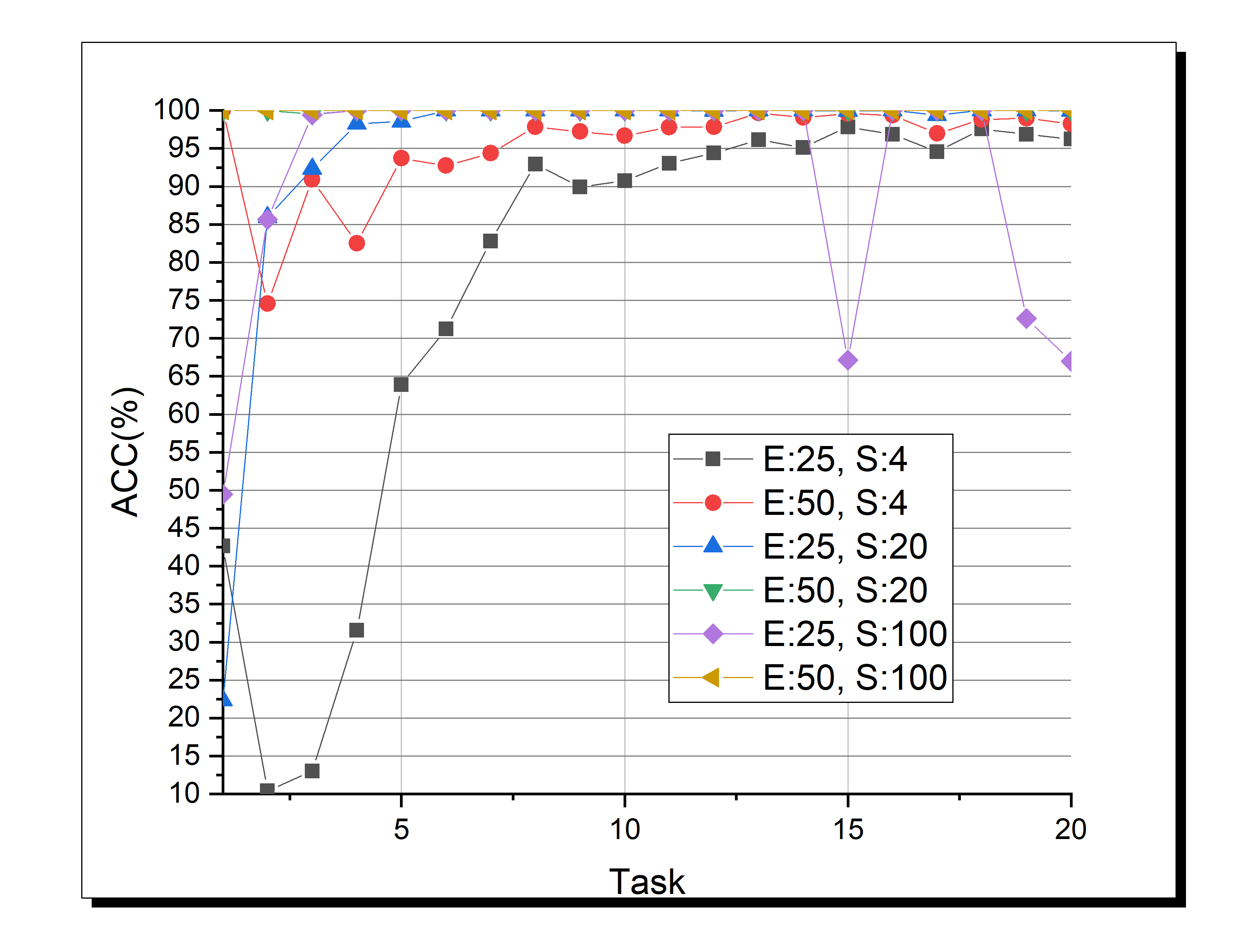}
    \caption{Results for the 20-Split miniImageNet data. E = Number of epochs the model gets trained, S = Number of synthetic data used per class as a generative replay.}
    \label{figure:f3}
\end{figure}

    We divided the miniImageNet dataset into twenty tasks with five classes for each task. We compare our results with several baselines in Table \ref{table:t4}. HAT\cite{a21} as a regularization-based method with no replay data achieves ACC = 59.45. A-GEM\cite{a30} and ER-RES\cite{a32} use architecture with 25.6M parameters along with memory replay. They store thirteen images of size $84\times84\times3$ per class during continual training. Reference\cite{a37} used both architecture-based and memory-based approaches together and outperformed other algorithms, and it achieves ACC = 62.07. Our model beats all existing models at a large margin and earns ACC = 100 when we take 100 synthetic samples per class and train it for 50 epochs. The model gives ACC = 77.4 when we take only four generated data per class and train it for 25 epochs. Our model takes 2900 seconds to learn the 20-Split miniImageNet data when trained for 50 epochs and used four samples generated per class. The ACCs for each task are given in Figure \ref{figure:f3} using a different combination between the number of epochs the model gets trained and the number of samples used as a generative replay. 
\begin{table}[hbt!]
\centering
\begin{tabular}{ |p{1.4cm}|p{1.2cm}|p{1cm}|p{1.2cm}|p{0.8cm}|p{1cm}|}
 \hline
 Method & ACC(\%) & BWT(\%) & Arch(MB) & M Replay& G Replay\\
 \hline
 HAT$^{*}$\cite{a21}   & 59.45    & -0.04 &123.6 & - & -\\
 \hline
 PNN$^{**}$\cite{a22} & 58.96    & 0.00 & 588 & - & -\\
 \hline
 ER-RES$^{*}$\cite{a32} & 57.32 & -11.34 & 102.6 & $\checkmark$  & -\\
 \hline
 A-GEM$^{*}$\cite{a30} & 52.43    & -15.23 & 102.6 & $\checkmark$ & -\\
 \hline
 ORD-FT\cite{a37}& 28.76 & -64.23 & 37.6 & - & -\\
 \hline
 ACL\cite{a37}& 62.07 & 0 & 113.1 & $\checkmark$ & -\\
 \hline
 ours& 100(0.00)    & 6.6(0.2) & 16 & - & $\checkmark$\\

 \hline
\end{tabular}
\caption{Results on 20-Split miniImageNet data measuring ACC(\%), BWT(\%), and memory size(MB) for architecture. M Replay: memory replay or actual data stored and G replay: Generated synthetic data. BWT positive means good. ($^{*}$) denotes result is reproduced by\cite{a37}.($^{**}$) denotes result is obtained using the re-implementation setup by\cite{a21}. All results are averaged over 3 runs and standard deviation is given in parentheses.}
\label{table:t4}
\end{table}

\subsection{Performance on 20-Split CIFAR100 Dataset}
\begin{figure}[hbt!]
\centering
    \includegraphics[width = 8cm, height = 5cm]{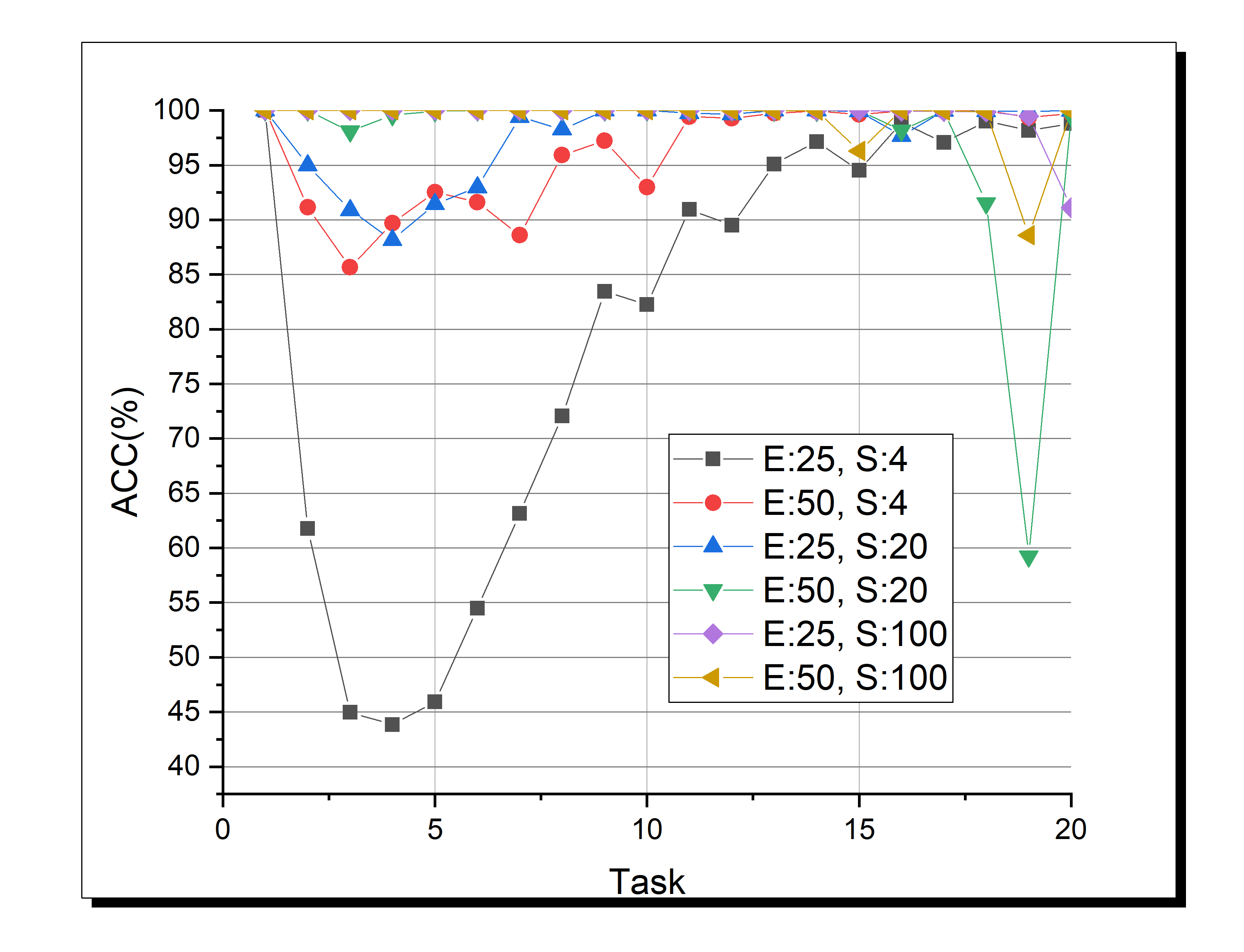}
    \caption{ Results for the 20-Split CIFAR100 data. E = Number of epochs the model gets trained, S = Number of synthetic data used per class as a generative replay.}
    \label{figure:f4}
\end{figure}

\begin{table}[hbt!]
\centering
\begin{tabular}{ |p{1.3cm}|p{1.4cm}|p{1cm}|p{1.2cm}|p{0.8cm}|p{0.8cm}|}

 \hline
 Method & ACC(\%) & BWT(\%) & Arch(MB) & M Replay& G Replay\\
 \hline
 HAT$^{*}$\cite{a21}   & 76.96    & 0.01 &27.2 & - & -\\
 \hline
 PNN$^{**}$\cite{a22} & 75.25   & 0.00 & 93.51 & - & -\\
 \hline
 ER-RES$^{o}$\cite{a32} & 54.38 & -21.99 & 25.4 & $\checkmark$  & -\\
 \hline
 A-GEM$^{o}$\cite{a30} & 66.78    & -15.09 & 25.4 & $\checkmark$ & -\\
 \hline
 ORD-FT\cite{a37}& 34.71 & -48.56 & 27.2 & - & -\\
 \hline
 ACL\cite{a37}& 78.08 & 0 & 25.1 & $\checkmark$ & -\\
 \hline
 ours& 99.52(0.59)    & 8.5(0.5) & 53 & - & $\checkmark$\\

 \hline
\end{tabular}
\caption{Results on 20-Split CIFAR100 data measuring ACC(\%), BWT(\%), and memory size(MB) for architecture. M Replay: memory replay or actual data stored and G replay: Generated synthetic data. BWT positive means good. ($^{*}$) denotes result is obtained by\cite{a37} using original provided code. ($^{**}$) denotes result is obtained using the re-implementation setup by\cite{a21}. ($^{o}$) denotes result is reported by\cite{a32}. All results are averaged over 3 runs and standard deviation is given in parentheses.}
\label{table:t5}
\end{table}
    We split the whole dataset into 20 tasks, where each contains five classes. We compare our results with other methods in Table \ref{table:t5}. ACL\cite{a37} is the most competitive baseline, takes 24.5 MB to save its architecture, takes 13 images per class(1300 images of size ($32\times32\times3$) in total) that require 16 MB of memory and achieves ACC = 78.08. HAT\cite{a21} does not depend on replay example but needs 27.2 MB for its architecture in which learns task-based attention mask reaching ACC = 76.96. PNN\cite{a22} guarantees zero forgetting. Our model outperforms all the existing models in a large margin and achieves ACC = 99.52 when we train the model for 50 epochs and use 100 synthetic data per class. The model gives ACC = 80.6 if we train it for 25 epochs with only four synthetic samples per class. Our model learns 20-Split CIFAR100 data in 1540 seconds if trained for 50 epochs and used only four synthetic samples per class as a generative replay. The ACCs for each task are given in Figure \ref{figure:f4} using a different combination between the number of epochs the model gets trained and the number of samples used as a generative replay.

\subsection{Performance on Permuted MNIST Dataset}
\begin{figure}[hbt!]
\centering
    \includegraphics[width = 8cm, height = 5cm]{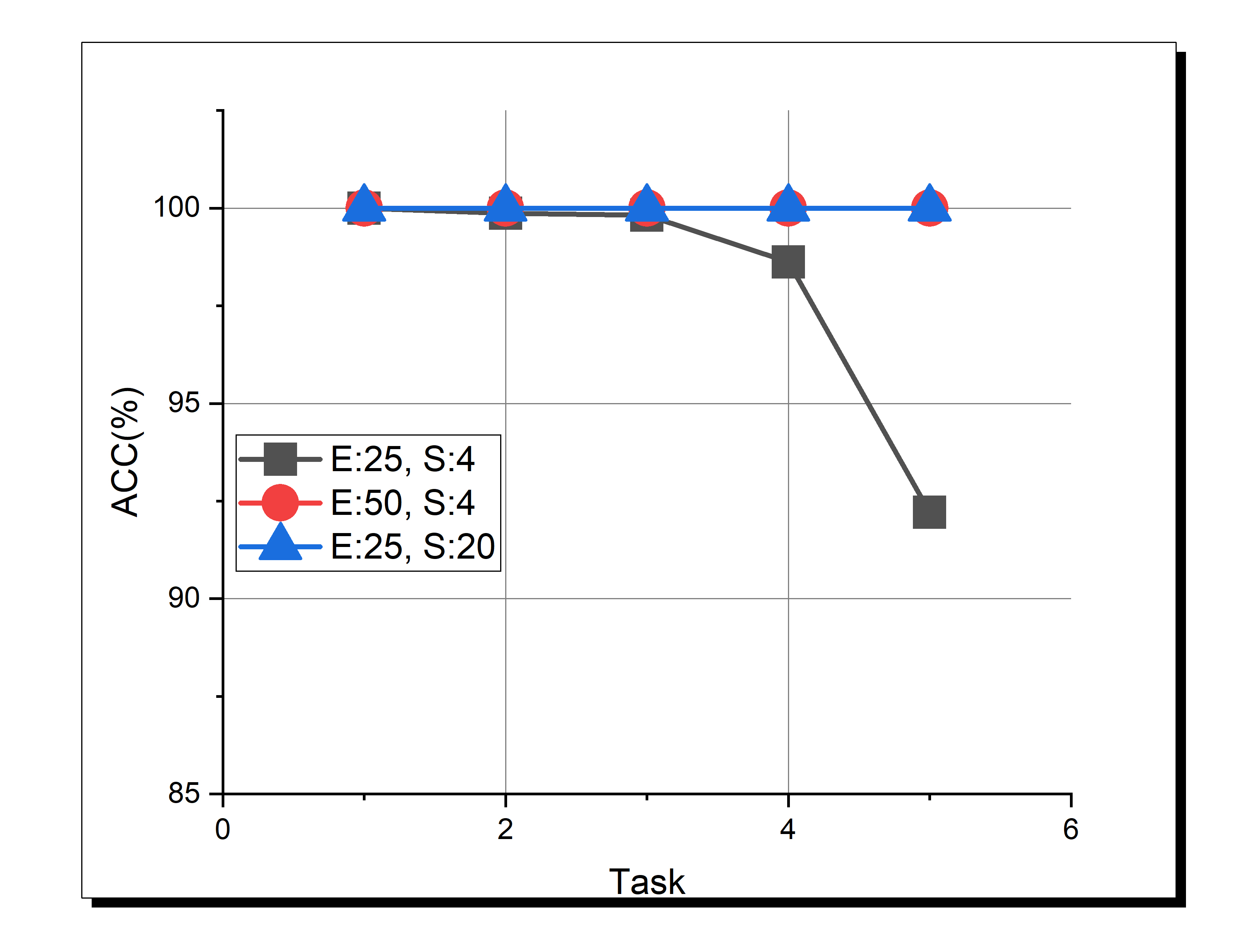}
    \caption{Results for the Permuted MNIST data. E = Number of epochs the model gets trained, S = Number of synthetic data used per class as a generative replay. For other((E:25, S:100), (E:50, S:20), and (E:50, S:100)) combinations ACC = 100\% for all tasks, so we did not plot.}
    \label{figure:f5}
\end{figure}

\begin{table}[hbt!]

\begin{tabular}{ |p{1.7cm}|p{2cm}|p{2cm}|p{1.3cm}|}
 \hline
 Method & ACC(\%) & BWT(\%) & Arch(MB)\\
 \hline
 EWC$^{oo}$\cite{a23}   & 88.2    & - &1.1\\
 \hline 
 HAT$^{o}$\cite{a21}   & 97.4    & - &2.8\\
 \hline
 UCB$^{o}$\cite{a41} & 91.44(0.04)   & -0.38(0.02) & 2.2\\
 \hline
 VCL$^{*}$\cite{a43} & 88.80(0.23) & -7.90(0.23) & 1.1\\
 \hline
 VCL-C$^{*}$\cite{a43} & 95.79(0.10)    & -1.38(0.12) & 1.1\\
 \hline
 PNN$^{**}$\cite{a22} & 93.5(0.07) & zero & N/A\\
 \hline
 ORD-FT\cite{a37}   & 44.91(6.61) & -53.69(1.91) & 1.1\\
 \hline
 ACL\cite{a37}& 98.03(0.01) & -0.01(0.01) & 2.4\\
 \hline
 ours& 100(0.00)    & 0.00(0.00) & 0.8\\

 \hline
\end{tabular}
\caption{Results on Permuted MNIST data measuring ACC(\%), BWT(\%), and memory size(MB) for architecture. BWT positive means good. ($^{oo}$) denotes result is reported by\cite{a21}. ($^{o}$) denotes result is reported by original work. ($^{*}$) denotes result is obtained by\cite{a37} using original code. ($^{**}$) denotes result is reported by\cite{a32}. All results are averaged over 3 runs and standard deviation is given in parentheses.}
\label{table:t6}
\end{table}
    One of the popular variants of the MNIST dataset in continual learning is Permuted MNIST. We divide the dataset into five tasks where each contains two classes. We compare our approach with other methods in Table \ref{table:t6}. HAT\cite{a21} achieves ACC = 91.6 using an architecture of size 1.1 MB. Vanilla VCL\cite{a43} improves ACC and BWT by 7 and 6.5, respectively, using k-means core-set-memory size of 200 samples per task(6.3 MB) and architecture size of 1.1 MB. PNN\cite{a22} achieves ACC = 93.5 with zero forgetting. ACL\cite{a37} achieves ACC = 98.03 takes 0.2 MB for adding 55k parameters for each task, and occupies a total of 2.5 MB memory. Our model outperforms all methods and achieves ACC = 100 when we train it for 50 epochs with four synthetic samples per class as a generative replay and occupies 0.8 MB for its architecture. If we train the model for 25 epochs with only four synthetic samples per class, it gives ACC = 98.12. The model learns Permuted MNIST dataset in 1300 seconds when trained for 50 epochs with same number of synthetic data.  The ACCs for each task are given in Figure \ref{figure:f5} using a different combination between the number of epochs the model gets trained and the number of samples used as a generative replay.  

\subsection{Performance on 5-Split MNIST Dataset}
\begin{figure}[hbt!]
    \includegraphics[width = 8cm, height = 5cm]{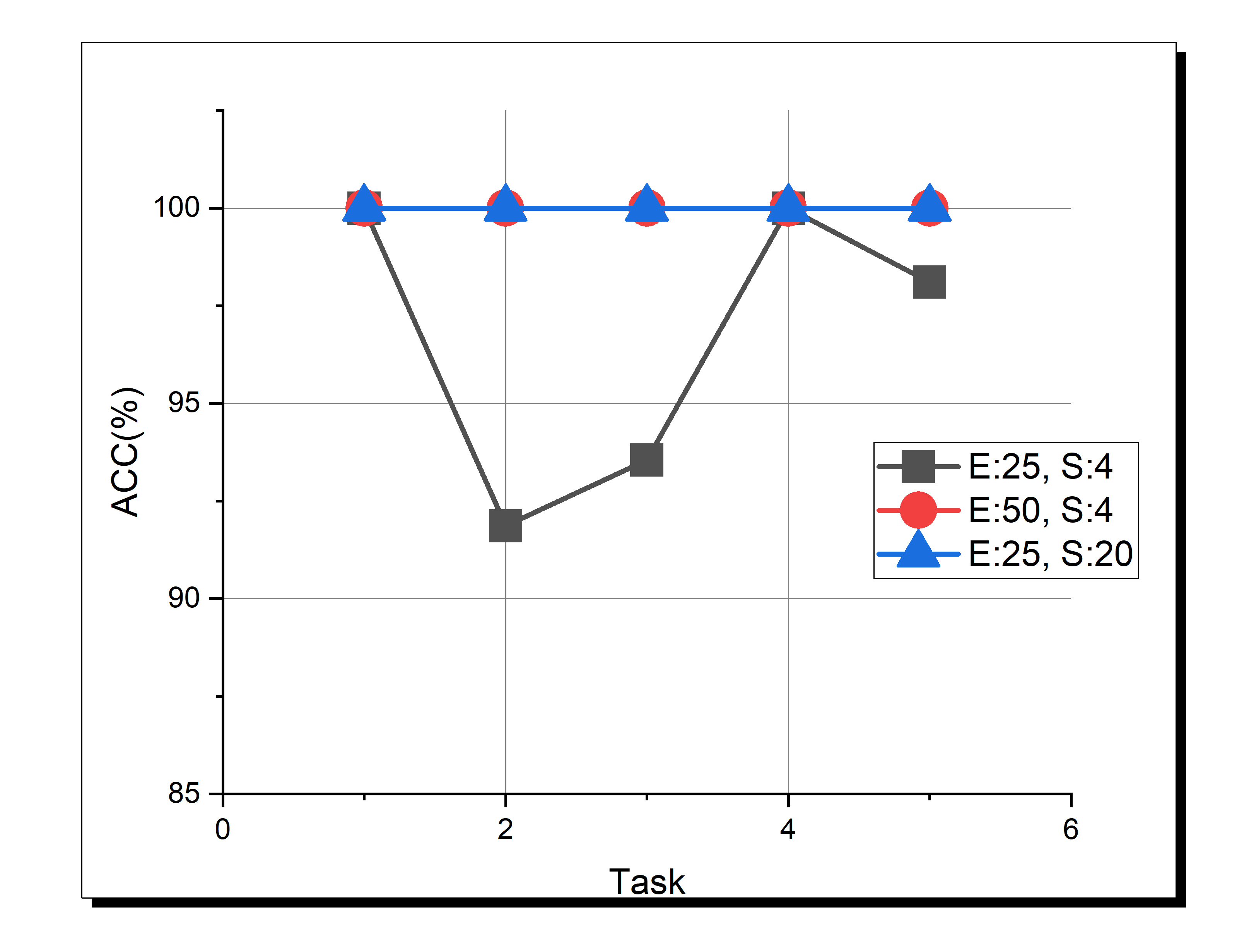}
    \caption{Results for the MNIST data. E = Number of epochs the model gets trained, S = Number of synthetic data used per class as a generative replay. For other((E:25, S:100), (E:50, S:20), and (E:50, S:100)) combinations ACC = 100\% for all tasks, so we did not plot.}
    \label{figure:f6}
\end{figure}

\begin{table}[hbt!]

\begin{tabular}{ |p{1.7cm}|p{2cm}|p{2cm}|p{1.3cm}|}
 \hline
 Method & ACC(\%) & BWT(\%) & Arch(MB)\\
 \hline
 EWC$^{oo}$\cite{a23}   & 95.78(0.35)    & -4.2(0.21) &1.1\\
 \hline 
 HAT$^{oo}$\cite{a21}   & 99.59(0.01)    & 0.00(0.04) &1.1\\
 \hline
 UCB$^{o}$\cite{a41} & 99.63(0.02)   & 0.00(0.00) & 2.2\\
 \hline
 VCL$^{*}$\cite{a43} & 95.97(1.03) & -4.62(1.28) & 1.1\\
 \hline
 VCL-C$^{*}$\cite{a43} & 93.6(0.20)    & -3.10(0.20) & 1.7\\
 \hline
 GEM$^{*}$\cite{a29} & 94.34(0.82) & -2.01(0.05) & 6.5\\
 \hline
 ORD-FT\cite{a37}   & 65.96(3.53) & -40.15(4.27) & 1.1\\
 \hline
 ACL\cite{a37}& 99.76(0.03) & 0.01(0.01) & 1.6\\
 \hline
 ours& 100(0.00)    & 0.00(0.00) & 0.8\\

 \hline
\end{tabular}
\caption{Results on 5-Split MNIST data measuring ACC(\%), BWT(\%), and memory size(MB) for architecture. BWT positive means good. ($^{oo}$) denotes result is reported by \cite{a41}. ($^{o}$) denotes result is taken from original work. ($^{*}$) denotes result is obtained by \cite{a37} using original provided code. All results are averaged over 3 runs and standard deviation is given in parentheses.}
\label{table:t7}
\end{table}
    We divide the MNIST dataset into five tasks, where each consists of two classes. We compare our results with other existing models in Table \ref{table:t7}. EWC, HAT, UCB, and Vanilla VCL are regularization-based methods with no memory replay are provided in that Table. Methods relying on memory only(GEM) and VCL with k-means core-set(VCL-C) where 40 samples are stored per task. ACL gives ACC = 99.76 with zero-forgetting outperforming UCB with ACC = 99.63 which uses 40\% more memory. ACL uses only architecture growth(no experience replay), where 54.3k private parameters are added for each task resulting in a memory requirement of 1.6 MB for all private modules. ACL's architecture has a total of 420.1k parameters. Our method outperforms all existing models, achieves ACC = 100 with BWT = 0 when we train it for 50 epochs with only four synthetic data per class as a generative replay, and learns the MNIST dataset in 660 seconds.  The ACCs for each task are given in Figure \ref{figure:f6} using a different combination between the number of epochs the model gets trained and the number of samples used as a generative replay.

\subsection{Performance on SVHN Dataset}
\begin{figure}[hbt!]
\centering
    \includegraphics[width = 8cm, height = 5cm]{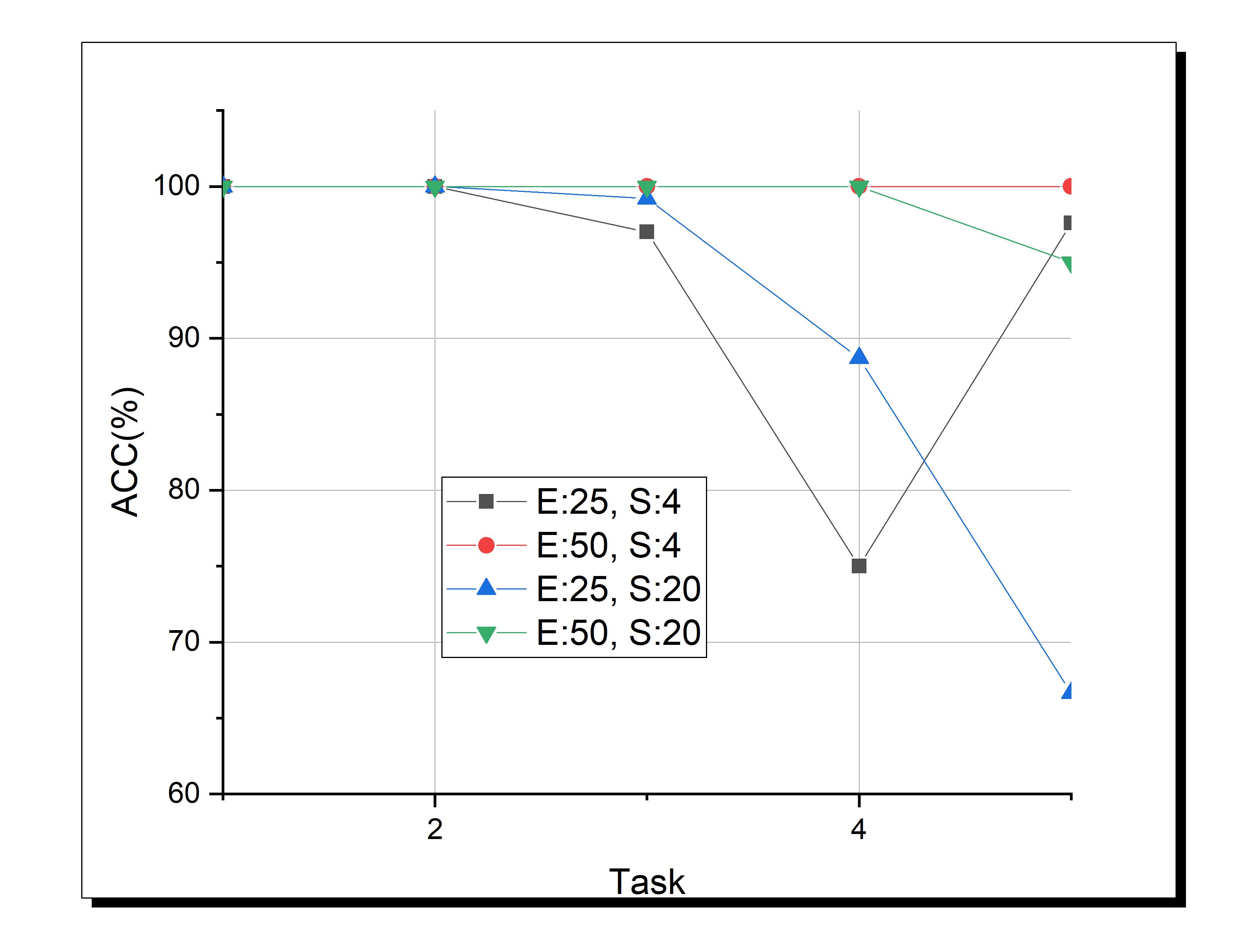}
    \caption{ Results for the SVHN data. E = Number of epochs the model gets trained, S = Number of synthetic data used per class as a generative replay.}
    \label{figure:f7}
\end{figure}
    We divide the SVHN dataset into five tasks, where each consists of two classes. Our model learns the whole dataset continually around 800 seconds when trained for 50 epochs and used only four synthetic samples per class as a generative replay. The network has a total of 3388428 parameters that occupies 13 MB of memory. The model gives ACC = 100. How the model performs in each task is given in Figure \ref{figure:f7} with few combinations between the number of epochs the model gets trained and the number of samples used as a generative replay. 

\subsection{Performance on Fashion-MNIST dataset}
\begin{figure}[hbt!]
\centering
    \includegraphics[width = 8cm, height = 5cm]{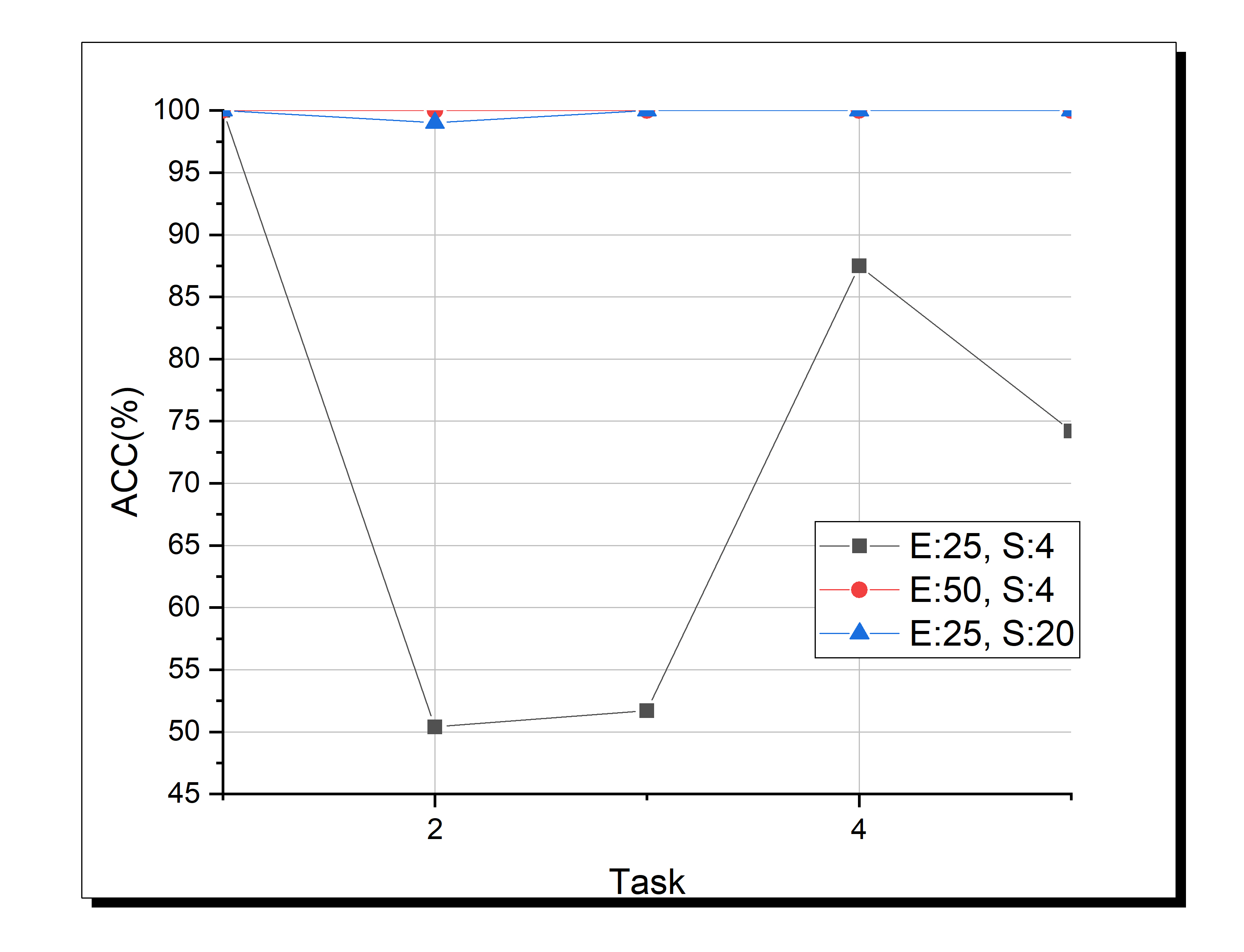}
    \caption{Results for the Fashion-MNIST data. E = Number of epochs the model gets trained, S = Number of synthetic data used per class as a generative replay.}
    \label{figure:f8}
\end{figure}

    We divide the Fashion-MNIST dataset into five tasks, where each consists of two classes. Our model gives ACC = 100, learns the whole dataset continually around 600 seconds when trained for 50 epochs, and used only four synthetic samples per class as a generative replay. The network has a total of 199019 parameters that occupies 0.8 MB of memory. How the model performs in each task is given in Figure \ref{figure:f8} with few combinations between the number of epochs the model gets trained and the number of samples used as a generative replay. 
\subsection{Performance on EMNIST Dataset}
\begin{figure}[hbt!]
\centering
    \includegraphics[width = 8cm, height = 5cm]{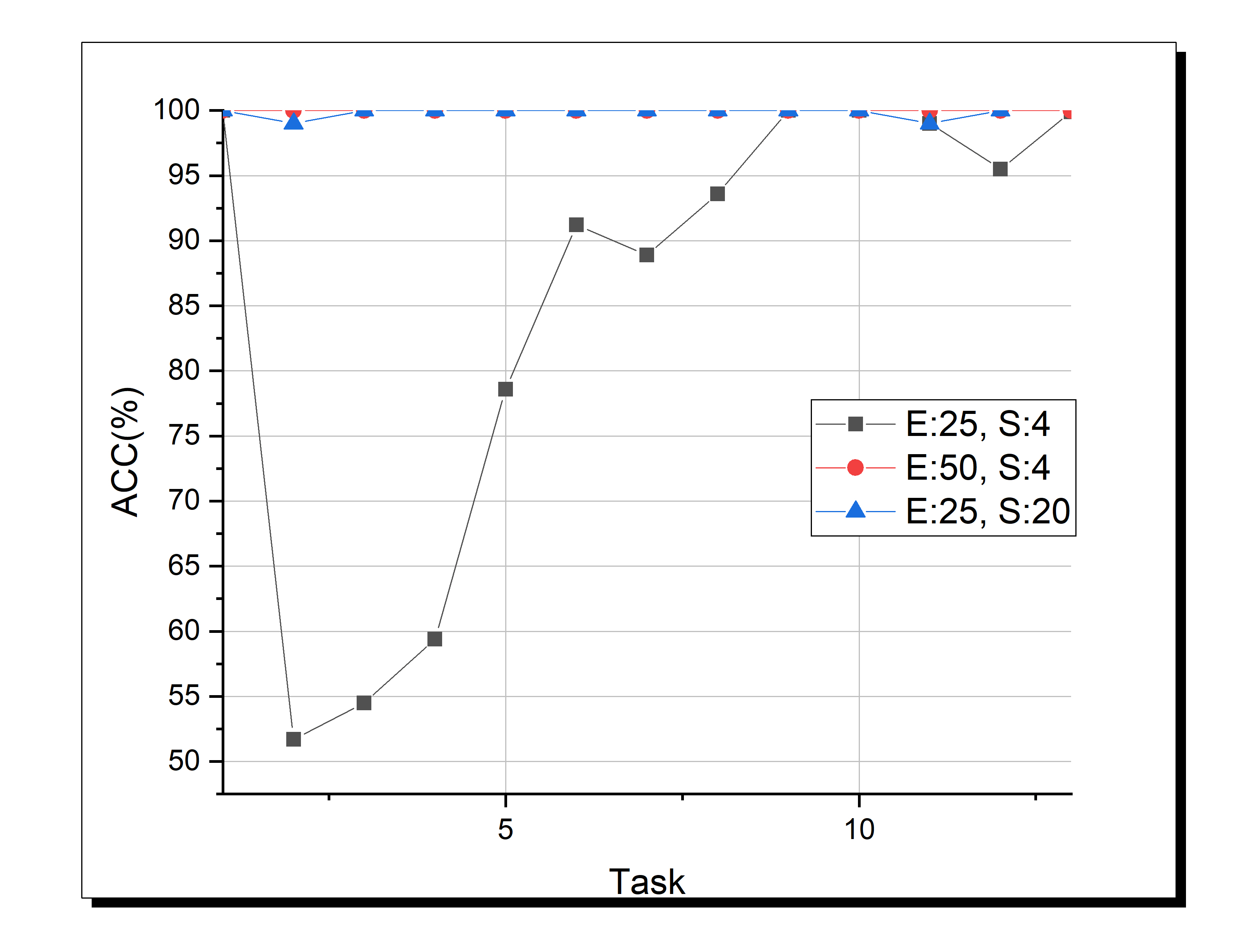}
    \caption{Results for the EMNIST data. E = Number of epochs the model gets trained, S = Number of synthetic data used per class as a generative replay.}
    \label{figure:f9}
\end{figure}
    We divide the EMNIST dataset into thirteen tasks, where each consists of two classes. Our model gives ACC = 100, learns the whole dataset continually around 1800 seconds when trained for 50 epochs, and used only four synthetic samples per class as a generative replay. The network has a total of 562939 parameters that occupies 2 MB of memory. The model's performance in each task is given in Figure \ref{figure:f9} with few combinations between the number of epochs the model gets trained and the number of samples used as a generative replay. 

\subsection{Performance on CIFAR10 Dataset}
\begin{figure}[hbt!]
\centering
    \includegraphics[width = 8cm, height = 5cm]{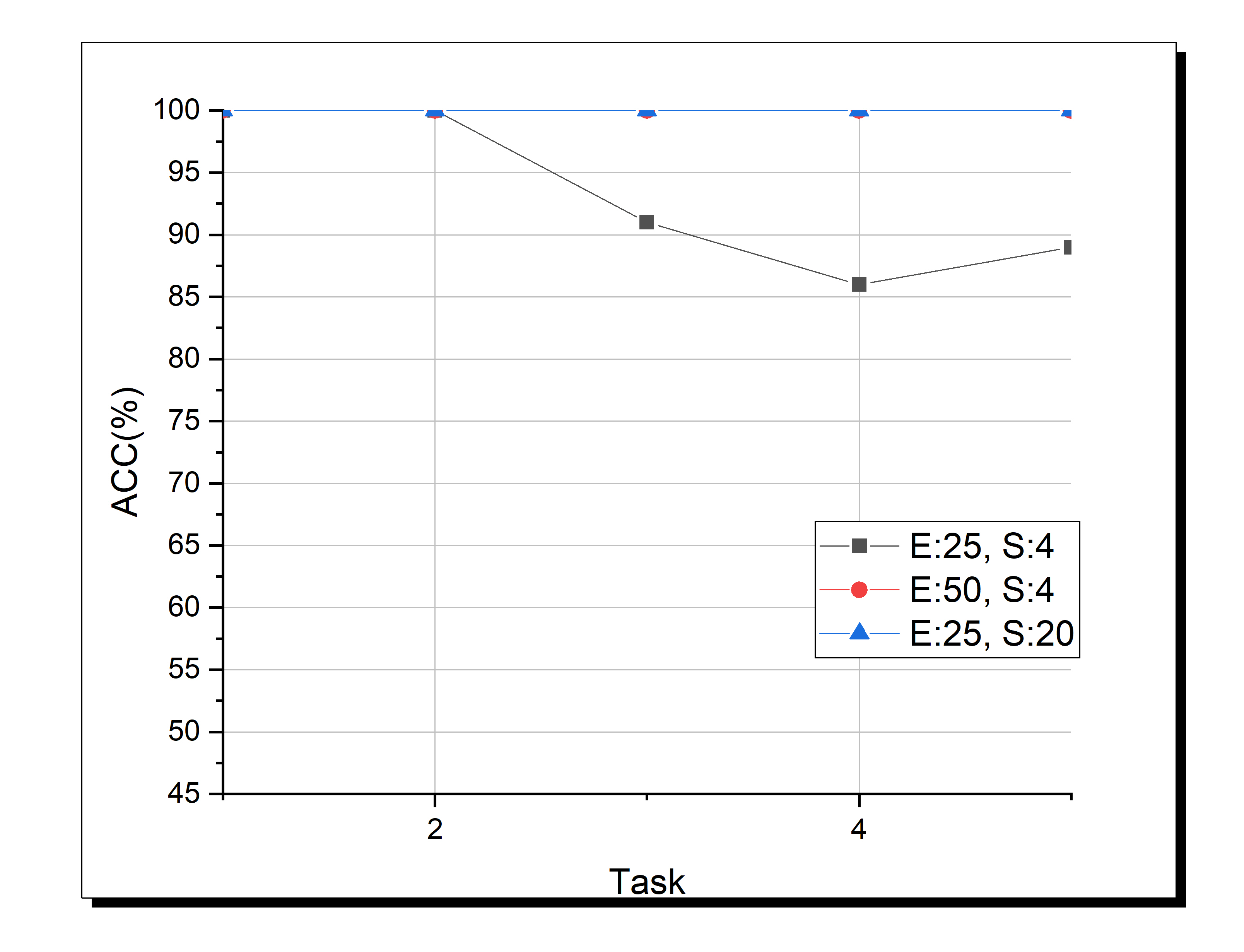}
    \caption{Results for the CIFAR10 data. E = Number of epochs the model gets trained, S = Number of synthetic data used per class as a generative replay.}
    \label{figure:f10}
\end{figure}
    We divide the CIFAR10 dataset into five tasks, where each consists of two classes. Our model gives ACC = 100, learns the whole dataset continually around 800 seconds when trained for 50 epochs, and used only four synthetic samples per class as a generative replay. The network has a total of 3388428 parameters that occupies 13 MB of memory. The model's performance in each task is given in Figure \ref{figure:f10} with few combinations between the number of epochs the model gets trained and the number of samples used as a generative replay. 

\section{Conclusion}
    In this paper, we propose a novel hybrid continual learning algorithm that grows in size with each task and factorizes the representation learned for a sequence of tasks into task-invariant and task-specific sub-spaces. The learning method combines generative replay and architecture-based approaches together. We show that it gives excellent performance on datasets having more classes or more tasks like miniImagenet and CIFAR100. We establish a new state-of-the-art on continual learning benchmark datasets. For future work, we are interested in extending this work in class incremental learning.
\clearpage
\newpage
{\small
\bibliographystyle{plain}
\bibliography{egbib}
}

\end{document}